\documentclass[twocolumn,prl,amsmath,amssymb,superscriptaddress]{revtex4-2}

\usepackage{graphicx}
\usepackage{hyperref}
\usepackage{booktabs}
\usepackage{siunitx}
\usepackage{xcolor}

\DeclareSIUnit{\sample}{sample}

\begin{document}

\title{Lead Zirconate Titanate Reservoir Computing for Classification of Written and Spoken Digits}

\author{Thomas Buckley}
\affiliation{Manning College of Information and Computer Science, University of Massachusetts Amherst, MA 01003}
\affiliation{Now at Harvard}

\author{Leslie Schumm}
\affiliation{Independent Researcher, Chicago, IL}

\author{Manor Askenazi}
\affiliation{Biomedical Hosting, Arlington, MA}

\author{Edward Rietman}
\affiliation{Manning College of Information and Computer Science, University of Massachusetts Amherst, MA 01003}
\affiliation{Small-Technology Incubator, Hanover, NH 03755}

\begin{abstract}
In this paper we extend our earlier work of (Rietman et al.\ 2022) presenting an application of physical Reservoir Computing (RC) to the classification of handwritten and spoken digits. We utilize an unpoled cube of Lead Zirconate Titanate (PZT) as a computational substrate to process these datasets. Our results demonstrate that the PZT reservoir achieves 89.0\% accuracy on MNIST handwritten digits, representing a 2.4 percentage point improvement over logistic regression baselines applied to the same preprocessed data. However, for the AudioMNIST spoken digits dataset, the reservoir system (88.2\% accuracy) performs equivalently to baseline methods (88.1\% accuracy), suggesting that reservoir computing provides the greatest benefits for classification tasks of intermediate difficulty where linear methods underperform but the problem remains learnable. PZT is a well-known material already used in semiconductor applications, presenting a low-power computational substrate that can be integrated with digital algorithms. Our findings indicate that physical reservoirs excel when the task difficulty exceeds the capability of simple linear classifiers but remains within the computational capacity of the reservoir dynamics.
\end{abstract}

\maketitle

\section{Introduction}
\label{sec:introduction}

\begin{figure*}[t]
\centering
\includegraphics[width=\textwidth]{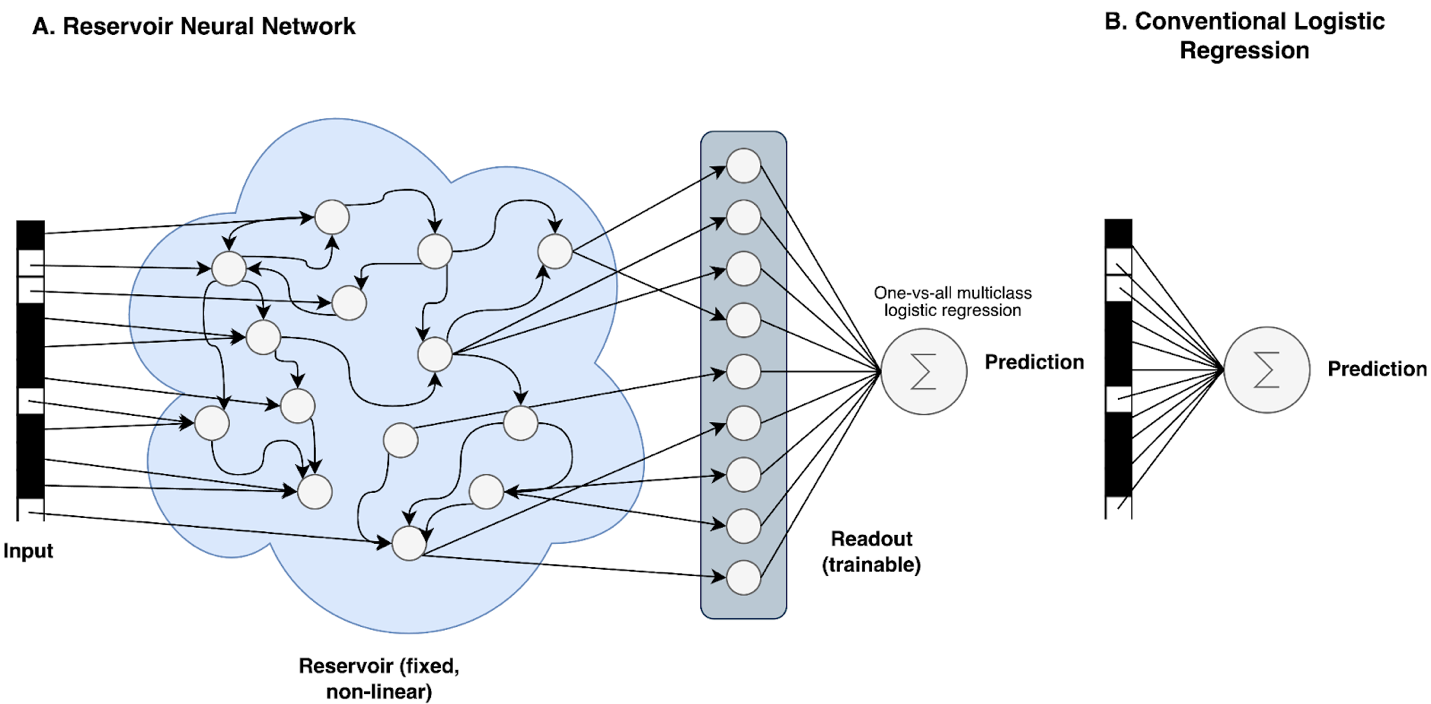}
\caption{(A) The physical reservoir neural network uses a fixed, non-linear device to map input signals to a high-dimensional temporal output. These higher-dimensional features are then classified by a very simple model such as a logistic regression. This regression is the trainable element in the reservoir system, reducing the effective computational cost of the entire network. The overall goal of the RC framework is to generate features that are highly separable. (B) We compare the physical reservoir to using only the regression model from the readout layer. This experiment measures the relative performance increase from using the reservoir.}
\label{fig:reservoir_concept}
\end{figure*}

Deep learning has become an incredibly useful technique for tasks ranging from natural language prediction to time-series classification~\cite{lecun2015deep}. However, these models often have a huge number of parameters making them costly to train using conventional techniques like backpropagation. Alternatively, it has been shown that a very large ``black box'' network with rich dynamics connected to a smaller trainable layer can also generate useful representations~\cite{maass1995computational}. Known as physical Reservoir Computing, this technique leverages the temporal expansion produced by stimulating a fixed, non-linear physical device to make input signals separable~\cite{cucchi2022hands,zhang2023survey} (Fig.~\ref{fig:reservoir_concept}). Some examples of physical reservoirs include the use of a water bucket and paddle for XOR classification~\cite{fernando2003pattern}, carbon nanotubes for modeling highly non-linear dynamical systems~\cite{dale2016evolving}, polymer fibers for iris classification~\cite{cucchi2022hands}, and chiral magnets for time series modeling~\cite{lee2024task}.

A critical question in reservoir computing research is identifying which tasks benefit most from physical reservoir implementations. While reservoirs are typically evaluated for their computational capacity or expressivity---the ability to approximate a greater number of functions~\cite{cucchi2022hands}---practical deployment requires understanding when a reservoir provides meaningful improvements over simpler computational approaches. In this study, we demonstrate that lead zirconate titanate (PZT), can produce an expressive reservoir for certain machine learning tasks while providing minimal benefit for others. This task-dependent performance suggests important insights about the applicability of physical reservoir computing.

Specifically, we use a small cubic block of unpoled PZT to transform input signals into a nonlinear temporal expansion. This material is well suited for reservoir computation because it exhibits both (1) nonlinearity, and (2) a fading memory, two conditions required for effective reservoir systems~\cite{cucchi2022hands}. Physical nonlinear effects have already been exploited in nonlinear I-V curves of memristors and transistors~\cite{cucchi2022hands,kulkarni2012memristor,du2017reservoir,appeltant2011information}. Here, we also leverage the nonlinear I-V curve of PZT~\cite{balke2008current,rietman2022machine}. This is important because if the device were linear, like a resistor for example, inputs that were not linearly separable would remain inseparable after the transformation~\cite{cucchi2022hands}.

Additionally, PZT has a fading memory, partly due to the stress-strain relationship resulting from the piezoelectric effect. This fading memory allows previous driving signals to interact with the current signal, behaving similarly to a recurrent neural network. Another property making PZT appealing is the ability to programmatically modify the reservoir. A previous study showed that by applying pulse trains along with the driving signal, the overall response of the PZT cube could be adjusted~\cite{rietman2022machine}. This makes PZT a flexible material that can adapt to different tasks.

In this paper, we use the cubic block of PZT as a reservoir to classify both (1) the well-known MNIST (Modified National Institute of Standards and Technology database) handwritten digits dataset~\cite{lecun1998mnist} and (2) a spoken digits dataset~\cite{becker2024audiomnist}. We present a novel architecture for using physical reservoirs in image-processing, with results that illuminate when physical reservoirs provide computational advantages. On MNIST handwritten digits, our system achieves 89.0\% accuracy compared to 86.6\% for the best regression baseline---a statistically significant improvement that demonstrates the reservoir's ability to generate useful nonlinear features. Conversely, on AudioMNIST spoken digits, the reservoir performs equivalently to baseline methods (88.2\% vs 88.1\%), suggesting the task may be too simple for the additional complexity of the reservoir to provide benefits.

While previous studies have shown the potential of simulated and physical reservoirs in classifying MNIST handwritten digits~\cite{rietman2022machine,lee2023handwritten,schaetti2016echo}, this work represents the first real-world implementation that demonstrably outperforms traditional regression techniques on the same preprocessed input data. We show that reservoir benefits are task-dependent: comparing our results to a CNN upper bound (97.4\% for handwritten digits, 90.1\% for spoken digits) reveals that reservoirs provide the greatest advantage when linear methods significantly underperform optimal solutions. For the spoken digits, where the baseline regression already achieves near-optimal performance, the reservoir offers no practical improvement. This insight suggests that physical reservoirs are most valuable for problems of intermediate difficulty---too complex for linear methods but still within the computational capacity of the reservoir dynamics. Shown as a proof-of-concept here, this makes the case that a wide range of materials possess similar expressive power for appropriately matched tasks.

\begin{figure*}[t]
\centering
\includegraphics[width=\textwidth]{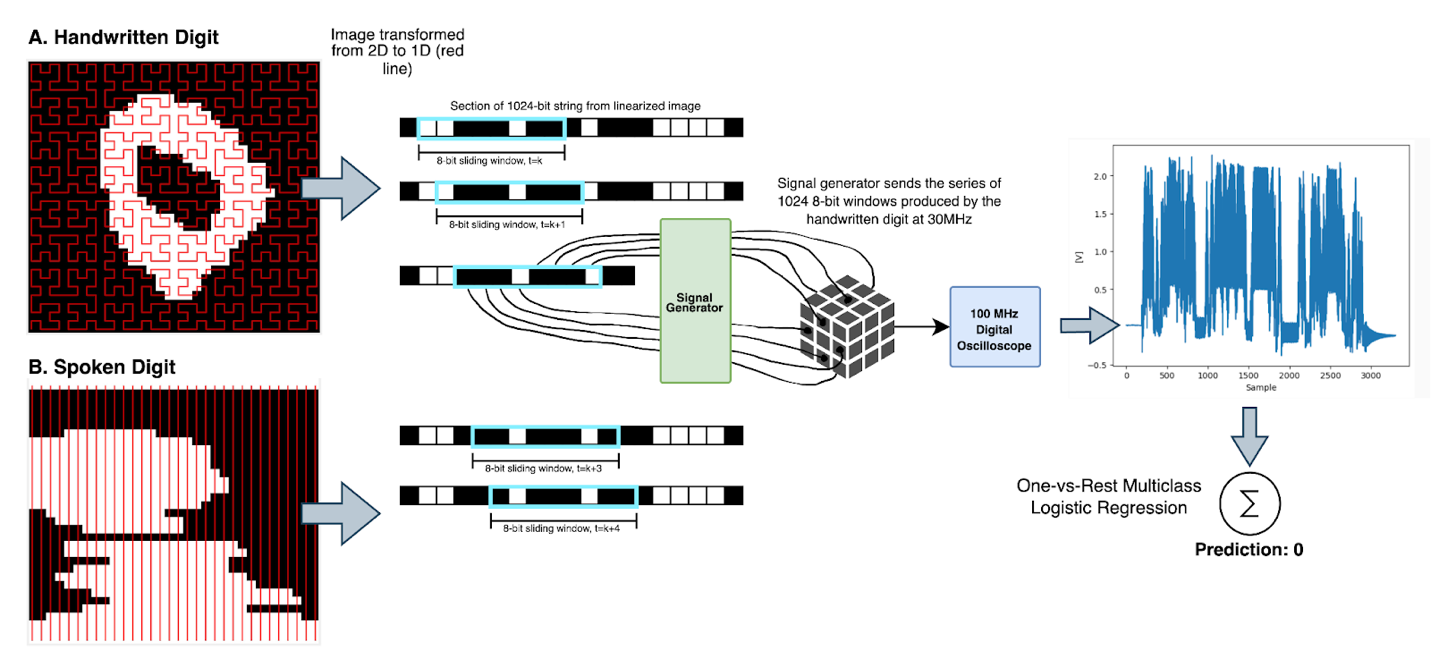}
\caption{The cube reservoir system architecture. In (A), a Hilbert curve is generated over the $32\times32$ binary handwritten digit, converting it to a 1D binary string. In (B), Mel-frequency cepstral coefficients (MFCCs) are computed for each spoken digit to produce $32\times32$ power spectrograms. These are then binarized by taking the mean value and converting values above the mean to 1 and values below to 0. This final binarized image (an example shown in the figure) is then scanned vertically to generate the 1D binary string (since the MFCCs are already a time series). An 8-bit sliding window with a stride of 1-bit is used to generate 1024 8-bit vectors, which are passed as a batch to the signal generator. The signal generator then applies the 8 binary values from each vector in parallel to a separate pad of the cube. This is done at high speed and recorded, producing the final image of the reservoir dynamics.}
\label{fig:architecture}
\end{figure*}

\section{Materials and Methods}
\label{sec:methods}

\subsection{The Cube System}
\label{sec:cube_system}

Our experimental system consists of four primary components: a custom high-speed arbitrary signal generator with 8 parallel outputs, a Lead Zirconate Titanate (PZT) cube serving as the physical reservoir, a high-speed oscilloscope for signal acquisition, and a digital readout layer for classification (Fig.~\ref{fig:architecture}). A photograph of the complete physical setup is shown in Fig.~\ref{fig:photo}.

The system operates by converting machine learning datasets into sequences of 8-bit binary values that are applied in parallel to the PZT cube. Each 8-bit value specifies whether each of the 8 input pads on the cube should be driven high (\SI{3.3}{\volt}) or low (\SI{0}{\volt}) at that time step. The oscilloscope measures the voltage response from a separate pad on the cube, capturing the complex temporal dynamics induced by the input signals. These captured voltage traces serve as high-dimensional feature representations that are then used to train a simple logistic regression classifier---the trainable readout layer of the reservoir system.

The key insight is that the reservoir performs a nonlinear temporal expansion: in our implementation, sequences of 1024 binary input values are transformed into floating-point vectors of up to 8{,}192 samples. This dimensional expansion, combined with the nonlinear dynamics of the PZT material, creates a feature space where the readout layer can more easily find separating hyperplanes for classification tasks. Any machine learning dataset that can be represented as a sequence of 8-bit binary values can theoretically be processed by this architecture---we demonstrate this flexibility by adapting both image data (handwritten digits) and audio data (spoken digits) to this format, as detailed in Sec.~\ref{sec:preprocessing} below.

\subsubsection{Signal Generator}

We implemented the signal generator using a Teensy 4.1 microcontroller (\SI{600}{\mega\hertz} ARM Cortex-M7 processor). The Teensy is programmed to receive datasets of up to 61{,}440 8-bit values via USB from a host PC. Each 8-bit value encodes the binary state (high or low) to be applied simultaneously to the 8 input pads of the PZT cube at a given time step.

Signal transmission to the cube is performed using Direct Memory Access (DMA) to achieve high-speed parallel output on a single 8-bit GPIO port. Through empirical testing, we determined the maximum reliable output frequency to be approximately \SI{40}{\mega\hertz}. To explore the effect of signal timing on reservoir performance, we varied the delay between successive 8-bit transmissions from \SI{5}{\nano\second} to \SI{50}{\nano\second} in our experiments. This timing parameter controls the rate at which input patterns are presented to the reservoir and consequently affects how the cube's temporal dynamics interact with the input sequence.

\subsubsection{The PZT Cube Reservoir}

\begin{figure}[t]
\centering
\includegraphics[width=\columnwidth]{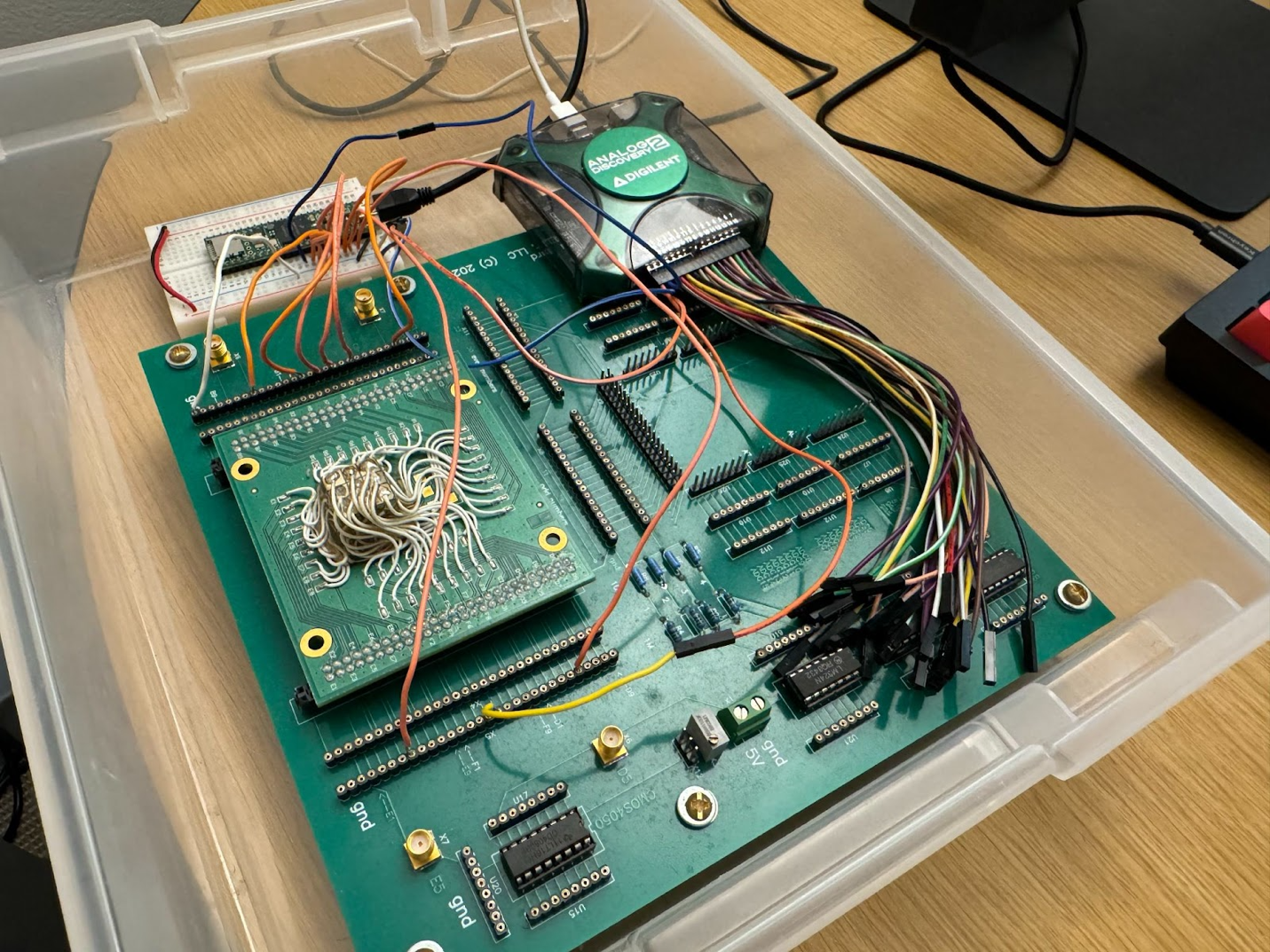}
\caption{Photograph of the cube system setup. The Teensy (top left) is used to generate a series of 8 parallel \SI{3.3}{\volt} signals at high speeds (\SI{30}{\mega\hertz}). The PZT cube (left middle) is used as the reservoir. The Analog Discovery 2 (top middle) is used to capture the reservoir dynamics at high speed (sampling frequency is \SI{100}{\mega\hertz}).}
\label{fig:photo}
\end{figure}

The physical reservoir consists of a cube fabricated from unpoled type 880 PZT material (American Piezo\textregistered, Inc.), measuring \SI{1.5}{\centi\meter} on each edge. This material was selected based on successful demonstration in previous reservoir computing work~\cite{rietman2022machine}. Each of the six cube faces was etched to create a $3\times3$ grid of electrically isolated contact pads, yielding 54 total pads across the cube surface. Electrical connections were established by silver-soldering thin wires to each pad, and the entire assembly was encapsulated in epoxy for mechanical stability and electrical insulation. A photograph of the fabricated PZT cube is shown in Fig.~\ref{fig:photo}.

PZT exhibits two critical properties for reservoir computing. First, the material displays strong nonlinearity in its current-voltage characteristics~\cite{balke2008current,rietman2022machine}, enabling the transformation of linearly inseparable inputs into separable feature representations. Second, PZT possesses a fading memory effect arising from its piezoelectric stress-strain dynamics: previous input signals create mechanical and electrical states that persist temporarily and interact with subsequent inputs. This temporal coupling mimics the recurrent connections in echo state networks and other reservoir computing architectures, allowing the system to process sequential information.

For this study, we used 8 pads as parallel inputs and monitored a single distinct pad as the output. The specific choice of input and output pads can affect reservoir performance, but systematic optimization of pad selection was beyond the scope of this demonstration.

\subsubsection{Signal Acquisition System}

We employed an Analog Discovery 2 USB oscilloscope (Digilent Inc.) to capture the reservoir's response to input signals. The oscilloscope was configured in a two-channel mode: one channel monitored a trigger signal from the Teensy, while the other channel recorded the analog voltage from the designated output pad of the PZT cube.

To synchronize data acquisition with the input signals, we programmed the oscilloscope to trigger on rising edges at \SI{1}{\volt} threshold. The trigger signal originates from an auxiliary GPIO pin on the Teensy that is set high at the start of each data sample transmission. We modified the Analog Discovery 2's onboard FPGA firmware to enable capture of longer signal traces---up to 16{,}384 samples at the maximum sampling rate of \SI{100}{\mega\sample\per\second}. However, due to the triggering mechanism's operation, only the second half of this buffer (8{,}192 samples) contains valid data corresponding to the reservoir response. This technical limitation constrains the maximum length of input sequences that can be processed in a single pass, though this rarely presents a practical bottleneck as longer sequences can be segmented or the sampling rate can be reduced to effectively capture extended temporal dynamics.

To ensure independence between classification samples, we implemented a 2-second reset period after processing each training or test example, during which all cube inputs are held at \SI{0}{\volt}. This allows transient states in the PZT material to dissipate, preventing carryover effects between different samples. However, within a single sample, temporal memory effects are explicitly desired---prior input values should influence the reservoir's response to subsequent values in the sequence, providing the recurrent dynamics essential for temporal processing.

\subsubsection{The Readout Layer}

For all experiments, including baseline comparisons, we used a standardized scikit-learn one-vs-rest logistic regression classifier as the readout layer. To ensure reproducibility and eliminate potential variations due to solver convergence differences, we fixed the \texttt{random\_state} parameter and set \texttt{max\_iter} to 1000 (higher than the default to accommodate the large feature spaces generated by the reservoir, which contain 8$\times$ more dimensions than the raw input). All other parameters were left at their default values. By maintaining identical readout layer architecture and hyperparameters across all experimental conditions---reservoir-based and baseline methods alike---we ensure that observed performance differences can be attributed solely to the feature representations rather than to classifier configuration.

\subsection{Dataset Preprocessing for Reservoir Computing}
\label{sec:preprocessing}

To evaluate our system on established machine learning benchmarks, we adapted two widely-used datasets: the MNIST handwritten digits dataset~\cite{lecun1998mnist} and the AudioMNIST spoken digits dataset~\cite{becker2024audiomnist}. Both datasets required conversion from their native formats (2D images and audio waveforms, respectively) into sequences of 8-bit binary values compatible with our hardware architecture. The preprocessing strategies differed between datasets to respect their inherent structure while conforming to the system's input requirements.

\subsubsection{MNIST Handwritten Digits}

We randomly sampled 5{,}000 examples from the full MNIST dataset to create a computationally manageable benchmark while maintaining representation across all digit classes. The preprocessing pipeline consisted of four steps:

\textit{Upsampling:} Each $28\times28$ grayscale image was upsampled to $32\times32$ pixels using nearest-neighbor interpolation.

\textit{Binarization:} Pixel values were converted to binary (0 or 1) based on a threshold, producing black-and-white digit representations.

\textit{Hilbert curve scanning:} Rather than a standard row-wise or column-wise raster scan, we applied a 2D Hilbert space-filling curve to traverse the $32\times32$ image. The Hilbert curve has superior locality-preserving properties compared to linear scans~\cite{moon2001analysis}---pixels that are close in 2D space tend to remain close in the resulting 1D sequence. This potentially allows the reservoir's temporal dynamics to better capture spatial structure in the images.

\textit{Sliding window encoding:} The resulting 1024-bit sequence was processed with an 8-bit sliding window using stride-1 advancement. At each position, the 8 bits within the window constitute one 8-bit value that will be applied to the 8 cube pads simultaneously. This produces 1{,}017 consecutive 8-bit values per image. However, for consistency across experiments and to accommodate system constraints, we standardized this to 1{,}024 8-bit values by appropriate padding or truncation.

This preprocessing necessarily reduces information content compared to the original grayscale images, as evidenced by the baseline regression accuracy of 86.6\% compared to CNN performance of 97.4\% on the original data. However, it enables direct comparison between reservoir-based and regression-based approaches on identical input representations.

\subsubsection{AudioMNIST Spoken Digits}

We randomly sampled 6{,}000 examples from the AudioMNIST dataset, which contains 30{,}000 recordings of spoken digits (0--9) from 60 speakers. The larger initial sample size (compared to handwritten digits) provided margin for potential sample removal during preprocessing while achieving approximately equal final dataset sizes.

The preprocessing pipeline consisted of five steps:

\textit{MFCC extraction:} For each audio recording, we computed Mel-frequency cepstral coefficients (MFCCs), a standard audio feature representation that captures spectral characteristics relevant to speech. We configured the MFCC extraction to use 32 mel-frequency bands and adjusted the hop length to produce 32 time frames, yielding $32\times32$ spectrograms.

\textit{Sample filtering:} Some audio recordings could not generate exactly $32\times32$ spectrograms due to duration constraints. These samples were removed from the dataset to maintain uniform dimensions.

\textit{Binarization:} The MFCC power values in each spectrogram were binarized by computing the mean value across all coefficients and assigning 1 to values above the mean and 0 to values below. This thresholding produced binary spectrograms while preserving the relative structure of spectral energy distribution.

\textit{Vertical scanning:} Unlike the handwritten digits, we used a simple vertical (column-wise) scan to convert the $32\times32$ binary spectrogram into a 1D sequence of 1{,}024 bits. The vertical scan was chosen instead of the Hilbert curve because the MFCC representation already encodes temporal information along one axis (time frames). Preserving this temporal ordering seemed more appropriate than imposing an alternative space-filling curve.

\textit{Sliding window encoding:} Identical to the handwritten digits pipeline, we applied an 8-bit sliding window with stride 1 to generate 1{,}024 8-bit values per spoken digit sample.

By applying similar preprocessing strategies to both visual and audio data, we demonstrate the generality of our reservoir computing architecture across different modalities.

\subsection{Experimental Design and Analysis}
\label{sec:experimental_design}

To rigorously assess whether the PZT cube provides computational benefits beyond simple linear transformation or noise injection, we designed a comprehensive comparison across multiple conditions.

\subsubsection{Reservoir System Evaluation}

We evaluated the reservoir system at three different signal timing delays (\SI{5}{\nano\second}, \SI{10}{\nano\second}, and \SI{20}{\nano\second} between successive 8-bit transmissions) to determine the optimal operating regime. For each timing condition, we processed all training examples through the cube, capturing the 8{,}192-sample voltage traces. These traces were then used to train the logistic regression readout layer. Performance was assessed on held-out test data (20\% of each dataset, using an 80/20 train-test split) processed through the same reservoir configuration.

\subsubsection{Baseline Comparisons}

To isolate the specific contribution of the PZT reservoir, we evaluated several baseline methods that use only the readout layer (logistic regression) applied directly to the preprocessed input data:

\textit{Direct logistic regression:} The simplest baseline applies logistic regression to the 1024-bit binary sequence (the same sequence that serves as input to the reservoir). This tests whether the reservoir adds value beyond the preprocessed representation itself.

\textit{Summed sliding window:} Since the reservoir processes 8 bits simultaneously at each time step, we created a baseline that mimics this windowing by summing values within each 8-bit window (producing integer values from 0 to 8) across the 1024-bit sequence. This tests whether simple aggregation of input features can explain reservoir performance.

\textit{Gaussian noise injection:} Physical reservoirs might simply benefit from adding noise to inputs, making them more distinguishable through stochastic separation. To test this hypothesis, we converted the binary input to floating-point values and added Gaussian noise ($\mu=0$, $\sigma=0.05$) before training the logistic regression.

By maintaining identical readout layer architecture across all conditions, any performance differences must arise from the input representations---either the raw/transformed inputs for baselines, or the reservoir-generated features for the cube system.

\subsubsection{CNN Upper Bound}

To establish a performance ceiling and assess how much room for improvement exists above baseline methods, we trained a convolutional neural network (CNN) on the original $32\times32$ binary versions of both datasets (i.e., after upsampling and binarization but before converting to 1D sequences). The CNN architecture (Table~\ref{tab:cnn}) represents a reasonable state-of-the-art approach for image classification, though it is not extensively optimized. This comparison reveals the cost of our 1D sequential encoding: the gap between CNN performance and baseline regression performance indicates how much information was lost in the conversion to sequences, and therefore represents the maximum potential improvement the reservoir could theoretically achieve.

\subsubsection{Statistical Analysis}

All experiments used 10-fold cross-validation to compute accuracy, F1 scores, and standard deviations, providing robust estimates of performance and variability. The fixed train-test split (80/20) was maintained within each fold to ensure consistency. We report both mean performance and standard deviation across folds, allowing assessment of statistical significance in performance differences between methods.

\begin{table}[t]
\centering
\caption{The CNN architecture used to classify the handwritten and spoken digits datasets, serving as a state-of-the-art comparison for the reservoir system.}
\label{tab:cnn}
\begin{tabular}{lc}
\toprule
Layer & Output Shape \\
\midrule
Input         & $(32, 32, 1)$ \\
Conv2D        & $(28, 28, 16)$ \\
MaxPooling2D  & $(14, 14, 16)$ \\
Conv2D        & $(10, 10, 32)$ \\
MaxPooling2D  & $(5, 5, 32)$ \\
Flatten       & $(800,)$ \\
Dense         & $(64,)$ \\
Dense         & $(10,)$ \\
\bottomrule
\end{tabular}
\end{table}

\begin{table*}[t]
\centering
\caption{Results using (A) the reservoir system, (B) regression techniques (i.e., using only the reservoir readout), or (C) a state-of-the-art CNN model on the MNIST Handwritten Digits dataset and the AudioMNIST Dataset. $K$-fold cross validation was used with 10 folds to compute accuracy, F1 scores, and standard deviations. We used 5{,}000 random samples from the Handwritten Digits dataset and 6{,}000 random samples from the Spoken Digits dataset. The time to generate cube responses is about 3 hours for each dataset. Performance of regression models are provided as realistic comparisons to the cube using the same input data---the 1D binarized version of either the handwritten digits or spoken digits. We used the scikit-learn one-vs-rest multiclass logistic regression models, with different inputs intended to approximate the effect of the reservoir.}
\label{tab:results}
\begin{tabular}{lcccc}
\toprule
 & \multicolumn{2}{c}{Handwritten Digits} & \multicolumn{2}{c}{Spoken Digits} \\
\cmidrule(lr){2-3} \cmidrule(lr){4-5}
Method & Test Accuracy (\%) & Test F1 (\%) & Test Accuracy (\%) & Test F1 (\%) \\
\midrule
\multicolumn{5}{l}{\textbf{A. Results from using Cube reservoir}} \\
Cube \SI{5}{\nano\second}   & 88.5 ($\sigma$=1.6)  & 88.2 ($\sigma$=1.6)  & 87.0 ($\sigma$=1.4)  & 87.2 ($\sigma$=1.2) \\
Cube \SI{10}{\nano\second}  & 89.0 ($\sigma$=1.1)  & 88.8 ($\sigma$=1.2)  & 88.2 ($\sigma$=0.7)  & 88.3 ($\sigma$=0.6) \\
Cube \SI{20}{\nano\second}  & 88.3 ($\sigma$=1.1)  & 88.0 ($\sigma$=1.2)  & 87.4 ($\sigma$=0.4)  & 87.6 ($\sigma$=0.3) \\
\midrule
\multicolumn{5}{l}{\textbf{B. Comparable methods on same dataset}} \\
Logistic regression                         & 86.2 ($\sigma$=1.2) & 85.9 ($\sigma$=1.3) & 88.1 ($\sigma$=1.0) & 88.2 ($\sigma$=0.9) \\
Log.\ reg.\ w/ summed windowed input        & 86.6 ($\sigma$=1.3) & 86.3 ($\sigma$=1.3) & 87.4 ($\sigma$=1.0) & 87.5 ($\sigma$=1.0) \\
Log.\ reg.\ w/ Gaussian noise               & 86.5 ($\sigma$=1.1) & 86.1 ($\sigma$=1.2) & 88.0 ($\sigma$=0.9) & 88.1 ($\sigma$=0.8) \\
\midrule
\multicolumn{5}{l}{\textbf{C. State-of-the-art method on original 2D version}} \\
CNN                                          & 97.4 ($\sigma$=0.4) & 97.4 ($\sigma$=0.4) & 90.1 ($\sigma$=1.7) & 91.1 ($\sigma$=1.6) \\
\bottomrule
\end{tabular}
\end{table*}

\section{Results}
\label{sec:results}

Our experimental evaluation compares the PZT reservoir system against multiple baseline methods across two distinct classification tasks: MNIST handwritten digits and AudioMNIST spoken digits. The central question is whether the physical reservoir provides computational advantages beyond simple linear processing of the same preprocessed input data. We structured our analysis to isolate the reservoir's contribution by maintaining identical readout layers (logistic regression) and preprocessing pipelines across all conditions, with only the feature representation varying between reservoir-based and baseline approaches.

\subsection{Overall Performance Summary}

Table~\ref{tab:results} presents comprehensive results across all experimental conditions. The data reveal a task-dependent pattern: the reservoir provides substantial benefits for handwritten digit classification but offers no meaningful advantage for spoken digit classification. This contrast, rather than representing an experimental failure, illuminates fundamental principles about when physical reservoir computing is most valuable.

\subsection{MNIST Handwritten Digits: Significant Reservoir Benefits}

For the handwritten digits task, the PZT reservoir demonstrates clear and statistically significant performance improvements over all baseline methods. At the optimal timing parameter of \SI{10}{\nano\second} delay, the reservoir system achieves 89.0\% accuracy ($\sigma$=1.1) and 88.8\% F1 score ($\sigma$=1.2) on the test set. This represents a 2.4 percentage point improvement in accuracy over the best-performing baseline method---logistic regression with summed windowed input, which achieves 86.6\% accuracy ($\sigma$=1.3).

\subsubsection{Timing Parameter Effects}

The reservoir's performance varies with the signal delay parameter, which controls how rapidly successive 8-bit patterns are presented to the cube. We tested three delay settings:

\begin{itemize}
\item \SI{5}{\nano\second} delay: 88.5\% accuracy ($\sigma$=1.6), 88.2\% F1 ($\sigma$=1.6)
\item \SI{10}{\nano\second} delay: 89.0\% accuracy ($\sigma$=1.1), 88.8\% F1 ($\sigma$=1.2) --- optimal
\item \SI{20}{\nano\second} delay: 88.3\% accuracy ($\sigma$=1.1), 88.0\% F1 ($\sigma$=1.2)
\end{itemize}

The \SI{10}{\nano\second} timing emerges as optimal, suggesting a ``sweet spot'' in the reservoir dynamics. At \SI{5}{\nano\second}, the signals may be too rapid for the PZT material's temporal dynamics to fully develop meaningful state differences between input patterns. At \SI{20}{\nano\second}, the slower presentation rate may allow too much decay of previous states, reducing the beneficial temporal coupling that enables sequential information processing. The \SI{10}{\nano\second} timing achieves an optimal balance where previous inputs sufficiently influence current states without overly rapid state transitions that the measurement system cannot adequately capture.

Additionally, the lower standard deviation at \SI{10}{\nano\second} ($\sigma$=1.1) compared to \SI{5}{\nano\second} ($\sigma$=1.6) indicates more consistent performance across cross-validation folds, suggesting that this timing parameter produces more stable reservoir dynamics.

\subsubsection{Comparison to Baseline Methods}

All three baseline regression approaches perform similarly to each other, clustering around 86.2--86.6\% accuracy:

\begin{itemize}
\item Direct logistic regression: 86.2\% accuracy ($\sigma$=1.2), 85.9\% F1 ($\sigma$=1.3)
\item Logistic regression with summed windowed input: 86.6\% accuracy ($\sigma$=1.3), 86.3\% F1 ($\sigma$=1.3)
\item Logistic regression with Gaussian noise: 86.5\% accuracy ($\sigma$=1.1), 86.1\% F1 ($\sigma$=1.2)
\end{itemize}

The near-identical performance across these baselines is highly informative. The summed windowing approach---designed to mimic the temporal aggregation of the reservoir's 8-bit parallel inputs---provides only a marginal 0.4 percentage point improvement over direct regression. This suggests that simple linear combinations of input features cannot explain the reservoir's benefits. Similarly, the Gaussian noise injection produces no meaningful improvement (86.5\% vs 86.2\% for direct regression), demonstrating that the reservoir's advantages cannot be attributed to stochastic separation through random noise addition. The PZT cube must therefore be performing genuinely nonlinear transformations that create more separable feature representations.

The reservoir's consistent improvement of approximately 2.4--2.8 percentage points across all baseline comparisons (89.0\% vs 86.2--86.6\%) demonstrates robust benefits from the physical transformation. This improvement is both statistically significant (given the standard deviations) and practically meaningful for deployment scenarios.

\subsubsection{Dimensional Expansion and Feature Quality}

The reservoir transforms the 1024-bit binary input sequences into floating-point vectors of 8{,}192 samples (captured from the oscilloscope), representing a 3--4$\times$ increase in dimensionality when accounting for the information content. This dimensional expansion alone does not guarantee improved performance---random projections to higher dimensions would not necessarily improve logistic regression performance. The key is that the reservoir generates nonlinear features through the PZT material's complex dynamics, creating representations where the digit classes are more linearly separable than in the original space.

\subsection{AudioMNIST Spoken Digits: Equivalent Performance}

In contrast to the handwritten digits results, the reservoir system provides no practical advantage for the spoken digits classification task. At the optimal \SI{10}{\nano\second} delay, the reservoir achieves 88.2\% accuracy ($\sigma$=0.7) and 88.3\% F1 ($\sigma$=0.6), which is statistically indistinguishable from the best baseline method---direct logistic regression at 88.1\% accuracy ($\sigma$=1.0) and 88.2\% F1 ($\sigma$=0.9).

\subsubsection{Timing Parameter Effects}

The reservoir's performance across timing parameters for spoken digits shows similar trends to handwritten digits but with more compressed differences:

\begin{itemize}
\item \SI{5}{\nano\second} delay: 87.0\% accuracy ($\sigma$=1.4), 87.2\% F1 ($\sigma$=1.2)
\item \SI{10}{\nano\second} delay: 88.2\% accuracy ($\sigma$=0.7), 88.3\% F1 ($\sigma$=0.6) --- optimal
\item \SI{20}{\nano\second} delay: 87.4\% accuracy ($\sigma$=0.4), 87.6\% F1 ($\sigma$=0.3)
\end{itemize}

Again, \SI{10}{\nano\second} emerges as optimal, with notably lower standard deviation ($\sigma$=0.7) compared to \SI{5}{\nano\second} ($\sigma$=1.4), indicating more stable performance. However, the absolute performance differences between timing conditions (1.2 percentage points from worst to best) are larger than for handwritten digits (0.7 percentage points), suggesting that the reservoir dynamics are more sensitive to timing parameters on this task despite providing no net accuracy benefit over baseline methods.

\subsubsection{Comparison to Baseline Methods}

All baseline approaches perform remarkably well on the spoken digits task:

\begin{itemize}
\item Direct logistic regression: 88.1\% accuracy ($\sigma$=1.0), 88.2\% F1 ($\sigma$=0.9)
\item Logistic regression with summed windowed input: 87.4\% accuracy ($\sigma$=1.0), 87.5\% F1 ($\sigma$=1.0)
\item Logistic regression with Gaussian noise: 88.0\% accuracy ($\sigma$=0.9), 88.1\% F1 ($\sigma$=0.8)
\end{itemize}

The direct logistic regression achieves essentially the same performance as the reservoir (88.1\% vs 88.2\%), indicating that the simple linear classifier can already effectively separate the spoken digit classes in the preprocessed MFCC-based representation. The summed windowing approach performs slightly worse (87.4\%), suggesting that for this particular dataset, the sliding window aggregation may actually discard useful information. The noise injection provides no benefit, confirming that stochastic effects are not responsible for classification performance.

One notable observation is that the reservoir shows slightly lower standard deviation ($\sigma$=0.7) compared to the best baseline ($\sigma$=1.0), suggesting marginally improved consistency across cross-validation folds. However, this small reduction in variance does not constitute a meaningful practical advantage and may simply reflect random variation.

\subsection{CNN Upper Bound Analysis: Interpreting Task Difficulty}

The CNN results provide crucial context for understanding why the reservoir helps with handwritten digits but not spoken digits. The CNN represents an approximate upper bound on achievable performance given our preprocessing ($32\times32$ binary representations).

\subsubsection{Handwritten Digits Performance Gap}

For handwritten digits, the CNN achieves 97.4\% accuracy ($\sigma$=0.4) and 97.4\% F1 ($\sigma$=0.4)---substantially higher than both the reservoir (89.0\%) and baseline methods (86.6\%). This creates an 10.8 percentage point gap between the best baseline method and the CNN upper bound. The reservoir captures approximately 22\% of this available improvement (2.4 out of 10.8 percentage points), demonstrating that significant additional performance gains are theoretically possible but require more sophisticated processing than our reservoir-plus-logistic-regression architecture provides.

The large gap between baseline and CNN performance (86.6\% vs 97.4\%) indicates that the handwritten digits task, even after preprocessing, contains substantial nonlinear structure that simple linear classifiers cannot exploit. The reservoir successfully captures some of this nonlinearity (reaching 89.0\%), but hierarchical feature learning in the CNN captures far more.

\subsubsection{Spoken Digits Performance Gap}

For spoken digits, the CNN achieves 90.1\% accuracy ($\sigma$=1.7) and 91.1\% F1 ($\sigma$=1.6)---only modestly higher than both the reservoir (88.2\%) and baseline methods (88.1\%). This creates just a 2.0 percentage point gap between the best baseline and the CNN upper bound.

The narrow gap between baseline and CNN performance (88.1\% vs 90.1\%) indicates that the spoken digits task, as preprocessed, is relatively simple---the logistic regression baseline has already extracted most of the learnable signal from the binarized MFCC features. With so little room for improvement, it is unsurprising that the reservoir cannot demonstrate benefits. The task is too easy for the additional complexity of physical reservoir computing to provide value.

\subsection{Task-Dependent Performance: A General Principle}

The contrasting results between handwritten and spoken digits reveal a fundamental principle: physical reservoirs provide the greatest computational benefits for tasks of intermediate difficulty. Specifically:

\begin{itemize}
\item When baseline linear methods perform poorly relative to optimal classifiers (handwritten digits: 86.6\% baseline vs 97.4\% CNN, creating an 10.8\,pp gap), there exists substantial nonlinear structure that reservoirs can partially exploit.
\item When baseline linear methods already approach optimal performance (spoken digits: 88.1\% baseline vs 90.1\% CNN, creating only a 2.0\,pp gap), the task lacks sufficient complexity for reservoirs to demonstrate advantages.
\end{itemize}

This pattern suggests that physical reservoir computing is most valuable for problems that are: (1) too complex for linear methods---problems with nonlinear structure that simple classifiers cannot capture; (2) within reservoir capacity---problems that don't require deep hierarchical feature learning beyond what temporal dynamics can provide; and (3) appropriately matched to the reservoir's dynamics---problems where the temporal or sequential structure aligns with the reservoir's natural processing capabilities.

The handwritten digits task falls into this ``Goldilocks zone'' of intermediate difficulty, while the spoken digits task is too simple to benefit from reservoir processing.

\section{Discussion}
\label{sec:discussion}

Our results demonstrate that a block of PZT can serve as an effective physical reservoir for classification tasks, but with an important caveat: the benefits are highly task-dependent. The contrasting outcomes between the MNIST handwritten digits and AudioMNIST spoken digits datasets reveal fundamental insights about when physical reservoir computing provides practical advantages.

\subsection{Performance Gains on Handwritten Digits}

For the MNIST handwritten digits dataset, the PZT reservoir achieves 89.0\% ($\sigma$=1.1) accuracy at the optimal \SI{10}{\nano\second} delay setting, representing a 2.4 percentage point improvement over the best baseline regression method at 86.6\% ($\sigma$=1.3) accuracy. This improvement is both statistically significant and practically meaningful, demonstrating that the reservoir successfully generates nonlinear features that enhance classification performance. The optimal timing parameter (\SI{10}{\nano\second} outperforming both \SI{5}{\nano\second} and \SI{20}{\nano\second}) suggests that the reservoir dynamics operate in a regime where signal interactions are neither too rapid to be captured nor too slow to maintain temporal coherence. Additionally, the reservoir expands the feature space by 3--4$\times$ compared to the raw input, projecting 1024 binary values to vectors of up to 8{,}192 floating-point values. This dimensional expansion, combined with the nonlinear transformation, enables the simple logistic regression readout layer to find decision boundaries that were inaccessible in the original feature space.

\subsection{Equivalent Performance on Spoken Digits}

In contrast, for the AudioMNIST spoken digits dataset, the reservoir system achieves 88.2\% ($\sigma$=0.7) accuracy, which is statistically indistinguishable from the baseline logistic regression at 88.1\% ($\sigma$=1.0) accuracy. The slightly lower standard deviation suggests improved consistency across trials, but this does not constitute a meaningful performance improvement. This null result is informative rather than disappointing---it reveals the limitations and appropriate application domain of physical reservoir computing.

\subsection{Understanding Task-Dependent Performance Through the CNN Upper Bound}

The CNN results provide crucial context for interpreting these findings. For handwritten digits, the CNN achieves 97.4\% accuracy, leaving an 10.8 percentage point gap between the optimal classifier and the best baseline regression method. The reservoir captures approximately 22\% of this performance gap (2.4 out of 10.8 percentage points). In contrast, for spoken digits, the CNN achieves only 90.1\% accuracy, leaving just a 2.0 percentage point gap between optimal and baseline performance. The reservoir's inability to improve upon baseline here suggests that the logistic regression has already extracted most of the learnable signal from the preprocessed data.

This pattern suggests a general principle: physical reservoirs provide the greatest benefit for tasks of intermediate difficulty---specifically, tasks where linear methods underperform due to insufficient feature complexity, but where the problem structure remains within the computational capacity of the reservoir's dynamics. When baseline methods already approach optimal performance, as in the spoken digits case, the additional complexity of the reservoir provides no computational advantage. When tasks are extremely difficult and require hierarchical feature learning (as evidenced by the 97.4\% CNN performance), simple reservoir systems may capture only a fraction of the available improvement.

\subsection{Comparison to Previous Physical Reservoir Implementations}

Our work represents, to the best of our knowledge, the first physical reservoir implementation that demonstrably outperforms baseline regression techniques when both are evaluated on identical preprocessed input data. A recent paper using simulated memristor crossbar arrays reported 91.1\% recognition accuracy on MNIST~\cite{du2017reservoir}, but the physical implementation achieved only 88.1\% due to device variations---equivalent to their baseline readout layer performance at 88\%. Another evaluation of a simulated Skyrmion reservoir demonstrated 88.2\% accuracy~\cite{lee2023handwritten} but compared this to an echo state network achieving 79.3\%~\cite{schaetti2016echo}, rather than to a simple logistic regression baseline. Without comparing to regression on the same preprocessed data, it remains unclear whether these systems provide computational advantages beyond their trainable readout layers.

Our approach of comparing the reservoir to multiple regression baselines applied to identical input data provides a rigorous test of whether the physical device contributes meaningful computation. The fact that adding Gaussian noise to the baseline input ($\sigma$=0.05) did not improve performance (86.5\% accuracy) demonstrates that the reservoir's benefits cannot be attributed to simple noise injection. Similarly, the windowed sum approach (86.6\% accuracy) fails to match the reservoir's performance, indicating that the PZT cube generates genuinely nonlinear transformations rather than simple aggregations of input features.

\subsection{Implications for Physical Reservoir Computing}

These results have important implications for the field of physical reservoir computing. First, they establish that material-based computation can provide measurable advantages over conventional digital processing for appropriately matched tasks. Second, they highlight the importance of task selection---not all machine learning problems benefit equally from reservoir computing. Researchers developing physical reservoirs should evaluate their systems against rigorous baselines on the same preprocessed data and should explore tasks across a spectrum of difficulty to identify optimal application domains.

Third, our preprocessing approach---converting 2D images to 1D binary sequences using Hilbert curves for handwritten digits and vertical scans for spoken digit spectrograms---necessarily reduces the information content available to both the reservoir and baseline methods. The strong CNN performance (97.4\% on handwritten digits) indicates substantial room for improvement. Future work could explore richer signal encoding schemes that preserve more spatial or spectral structure, potentially allowing reservoirs to capture more of the performance gap.

\subsection{Future Directions}

This proof-of-concept system suggests several directions for future work. Ensembles of multiple PZT devices could provide greater nonlinearity and richer dynamics, potentially capturing more of the performance gap toward CNN-level accuracy. More sophisticated sampling techniques could increase the effective dimensionality of the readout layer. Alternative dataset encodings---such as converting image features to sine wave frequencies as demonstrated in other physical reservoir work~\cite{cucchi2022hands}---may better match the temporal dynamics of PZT. Additionally, while we used exclusively binary signals to maximize output voltage, analog input signals could provide finer-grained control over the reservoir state and potentially improve performance.

Exploring the tunability of PZT through pulse train modulation~\cite{rietman2022machine} could enable task-adaptive reservoirs that adjust their dynamics to match problem characteristics. Investigating which specific properties of the handwritten digits task made it amenable to reservoir processing---perhaps related to the sequential structure imposed by the Hilbert scan or the inherent complexity of handwritten character variations---could guide selection of future application domains.

An especially promising avenue for future exploration involves thin-film implementations of quasicrystalline perovskite materials, which could combine the computational advantages of reservoir computing with novel topological properties. Details of this proposed research direction are presented in Appendix~\ref{sec:appendix}, including considerations for film growth, substrate selection, and potential applications to topological quantum computing~\cite{amaral2022exploiting,dorini2021two,forster2013quasicrystalline,huran2021two,krupski2015structure,yoneda1998growth,yu2020piezoelectricity,zali2014xray}.

\section{Conclusion}
\label{sec:conclusion}

We have demonstrated that a block of PZT serves as an effective physical reservoir for difficult classification tasks while providing no advantage for simpler tasks. For MNIST handwritten digits, our system achieves meaningful performance improvements over regression baselines on identical preprocessed data, establishing PZT as a viable substrate for reservoir computing. For AudioMNIST spoken digits, the equivalent performance between reservoir and baseline methods reveals that physical reservoirs are not universally beneficial---they excel when linear methods are insufficient but the problem remains tractable. This task-dependent performance provides important guidance for practitioners: physical reservoir computing is most valuable for problems of intermediate difficulty where conventional linear methods underperform but the computational capacity of the reservoir dynamics remains sufficient to extract useful features. Our work represents an important step toward understanding when and how physical substrates can augment conventional digital computation.

\begin{acknowledgments}
\end{acknowledgments}

\appendix

\section{Quasicrystalline BaTiO\texorpdfstring{$_3$}{3} for Machine Learning Reservoir Computing}
\label{sec:appendix}

Perovskites represent one of the most versatile crystal classes known, spanning applications from insulators and semiconductors to superconducting and magnetic materials. They exhibit piezoelectric, ferroelectric, and various optical phenomena, making them ideal candidates for computational substrates. The general perovskite structure follows the chemical formula ABX$_3$. For example, CaTiO$_3$ crystallizes in the Pbnm space group (number 62), where ion A occupies the center of the lattice, ion B is at the corners, and X is in the center faces.

As described in Ref.~\cite{rietman2022machine}, PZT, a perovskite material, is essentially quantum matter---a new realization of something that has had technological uses for some decades. By exploiting the ``flexible'' lattice of PZT crystals (see Figures 7 and 8 in Ref.~\cite{rietman2022machine}), it is possible to control the phonons in the material with electric signals, as demonstrated in this paper. For exploiting the properties of this quantum matter it is not necessary to operate at low temperature. At different temperatures you will get different results, so we hold the temperature fixed at \SI{300}{\kelvin}.

Theoretical work has proposed that piezoelectric materials may exhibit discontinuous changes resulting in direct signatures of 2D topological quantum phase transitions~\cite{yu2020piezoelectricity}. Specifically, applying an electric field can result in a jump in the piezoelectric tensor, directly related to topological quantum phase transitions observable in real systems. This phenomenon suggests that quasiperiodic piezoelectric materials could offer unique advantages for neural computation.

Building upon the demonstrated success of lead zirconate titanate (PZT) as a reservoir neural network substrate, this appendix outlines a promising avenue for future research: the development of two-dimensional quasicrystalline BaTiO$_3$ thin films for machine learning applications. Recent literature~\cite{dorini2021two,forster2013quasicrystalline,huran2021two} has shown that BaTiO$_3$ films can form quasicrystal structures on Pt~(111) substrates. By depositing these films on insulating layers such as MgO (cubic lattice) or BaO (hexagonal lattice), we anticipate the creation of a novel computational substrate that combines the unique properties of quasicrystals with the piezoelectric characteristics of perovskite materials.

Several studies have reported the growth of epitaxial thin films of BaTiO$_3$ on Au~(111) and Pt~(111) surfaces. Ref.~\cite{forster2013quasicrystalline} demonstrated two techniques for growing quasicrystals on Pt~(111) substrates: RF sputtering of BaTiO$_3$ and molecular beam epitaxy by simultaneously depositing BaO, Ti, and oxygen. Their work, along with detailed theoretical analyses by Ref.~\cite{huran2021two} and thermodynamics studies by Ref.~\cite{dorini2021two}, provides a foundation for the proposed research.

Since Pt is conductive, the proposed work would utilize insulating substrates. The lattice spacing for Pt~(111) is approximately 2.32--2.29~\AA~\cite{krupski2015structure}, while cubic BaTiO$_3$ exhibits spacing between 3.993--4.028~\AA\ depending on calcination temperature~\cite{zali2014xray}. MgO, with a cubic lattice spacing of 4.212~\AA, and BaO, with hexagonal lattice parameters ($a = 2.69$~\AA, $b = 2.69$~\AA, $c = 4.38$~\AA), offer promising alternatives as insulating substrates. Ref.~\cite{yoneda1998growth} has already demonstrated the growth of ultrathin BaTiO$_3$ films on MgO, though BaO may provide a better lattice match.

\bibliography{references}

\begin{thebibliography}{25}%
\makeatletter
\providecommand \@ifxundefined [1]{%
 \@ifx{#1\undefined}
}%
\providecommand \@ifnum [1]{%
 \ifnum #1\expandafter \@firstoftwo
 \else \expandafter \@secondoftwo
 \fi
}%
\providecommand \@ifx [1]{%
 \ifx #1\expandafter \@firstoftwo
 \else \expandafter \@secondoftwo
 \fi
}%
\providecommand \natexlab [1]{#1}%
\providecommand \enquote  [1]{``#1''}%
\providecommand \bibnamefont  [1]{#1}%
\providecommand \bibfnamefont [1]{#1}%
\providecommand \citenamefont [1]{#1}%
\providecommand \href@noop [0]{\@secondoftwo}%
\providecommand \href [0]{\begingroup \@sanitize@url \@href}%
\providecommand \@href[1]{\@@startlink{#1}\@@href}%
\providecommand \@@href[1]{\endgroup#1\@@endlink}%
\providecommand \@sanitize@url [0]{\catcode `\\12\catcode `\$12\catcode `\&12\catcode `\#12\catcode `\^12\catcode `\_12\catcode `\%12\relax}%
\providecommand \@@startlink[1]{}%
\providecommand \@@endlink[0]{}%
\providecommand \url  [0]{\begingroup\@sanitize@url \@url }%
\providecommand \@url [1]{\endgroup\@href {#1}{\urlprefix }}%
\providecommand \urlprefix  [0]{URL }%
\providecommand \Eprint [0]{\href }%
\providecommand \doibase [0]{https://doi.org/}%
\providecommand \selectlanguage [0]{\@gobble}%
\providecommand \bibinfo  [0]{\@secondoftwo}%
\providecommand \bibfield  [0]{\@secondoftwo}%
\providecommand \translation [1]{[#1]}%
\providecommand \BibitemOpen [0]{}%
\providecommand \bibitemStop [0]{}%
\providecommand \bibitemNoStop [0]{.\EOS\space}%
\providecommand \EOS [0]{\spacefactor3000\relax}%
\providecommand \BibitemShut  [1]{\csname bibitem#1\endcsname}%
\let\auto@bib@innerbib\@empty
\bibitem [{\citenamefont {LeCun}\ \emph {et~al.}(2015)\citenamefont {LeCun}, \citenamefont {Bengio},\ and\ \citenamefont {Hinton}}]{lecun2015deep}%
  \BibitemOpen
  \bibfield  {author} {\bibinfo {author} {\bibfnamefont {Y.}~\bibnamefont {LeCun}}, \bibinfo {author} {\bibfnamefont {Y.}~\bibnamefont {Bengio}},\ and\ \bibinfo {author} {\bibfnamefont {G.}~\bibnamefont {Hinton}},\ }\bibfield  {title} {\bibinfo {title} {Deep learning},\ }\href@noop {} {\bibfield  {journal} {\bibinfo  {journal} {Nature}\ }\textbf {\bibinfo {volume} {521}},\ \bibinfo {pages} {436} (\bibinfo {year} {2015})}\BibitemShut {NoStop}%
\bibitem [{\citenamefont {Maass}(1995)}]{maass1995computational}%
  \BibitemOpen
  \bibfield  {author} {\bibinfo {author} {\bibfnamefont {W.}~\bibnamefont {Maass}},\ }\href@noop {} {\emph {\bibinfo {title} {On the Computational Power of Noisy Spiking Neurons}}}\ (\bibinfo  {publisher} {Institutes for Information Processing Graz},\ \bibinfo {year} {1995})\BibitemShut {NoStop}%
\bibitem [{\citenamefont {Cucchi}\ \emph {et~al.}(2022)\citenamefont {Cucchi}, \citenamefont {Abreu}, \citenamefont {Ciccone}, \citenamefont {Brunner},\ and\ \citenamefont {Kleemann}}]{cucchi2022hands}%
  \BibitemOpen
  \bibfield  {author} {\bibinfo {author} {\bibfnamefont {M.}~\bibnamefont {Cucchi}}, \bibinfo {author} {\bibfnamefont {S.}~\bibnamefont {Abreu}}, \bibinfo {author} {\bibfnamefont {G.}~\bibnamefont {Ciccone}}, \bibinfo {author} {\bibfnamefont {D.}~\bibnamefont {Brunner}},\ and\ \bibinfo {author} {\bibfnamefont {H.}~\bibnamefont {Kleemann}},\ }\bibfield  {title} {\bibinfo {title} {Hands-on reservoir computing: a tutorial for practical implementation},\ }\href@noop {} {\bibfield  {journal} {\bibinfo  {journal} {Neuromorphic Computing and Engineering}\ }\textbf {\bibinfo {volume} {2}},\ \bibinfo {pages} {032002} (\bibinfo {year} {2022})}\BibitemShut {NoStop}%
\bibitem [{\citenamefont {Zhang}\ and\ \citenamefont {Vargas}(2023)}]{zhang2023survey}%
  \BibitemOpen
  \bibfield  {author} {\bibinfo {author} {\bibfnamefont {H.}~\bibnamefont {Zhang}}\ and\ \bibinfo {author} {\bibfnamefont {D.~V.}\ \bibnamefont {Vargas}},\ }\bibfield  {title} {\bibinfo {title} {A survey on reservoir computing and its interdisciplinary applications beyond traditional machine learning},\ }\href@noop {} {\bibfield  {journal} {\bibinfo  {journal} {IEEE Access}\ }\textbf {\bibinfo {volume} {11}},\ \bibinfo {pages} {81033} (\bibinfo {year} {2023})}\BibitemShut {NoStop}%
\bibitem [{\citenamefont {Fernando}\ and\ \citenamefont {Sojakka}(2003)}]{fernando2003pattern}%
  \BibitemOpen
  \bibfield  {author} {\bibinfo {author} {\bibfnamefont {C.}~\bibnamefont {Fernando}}\ and\ \bibinfo {author} {\bibfnamefont {S.}~\bibnamefont {Sojakka}},\ }\bibfield  {title} {\bibinfo {title} {Pattern recognition in a bucket},\ }in\ \href@noop {} {\emph {\bibinfo {booktitle} {Advances in Artificial Life}}}\ (\bibinfo  {publisher} {Springer Berlin Heidelberg},\ \bibinfo {year} {2003})\ pp.\ \bibinfo {pages} {588--597}\BibitemShut {NoStop}%
\bibitem [{\citenamefont {Dale}\ \emph {et~al.}(2016)\citenamefont {Dale}, \citenamefont {Miller}, \citenamefont {Stepney},\ and\ \citenamefont {Trefzer}}]{dale2016evolving}%
  \BibitemOpen
  \bibfield  {author} {\bibinfo {author} {\bibfnamefont {M.}~\bibnamefont {Dale}}, \bibinfo {author} {\bibfnamefont {J.~F.}\ \bibnamefont {Miller}}, \bibinfo {author} {\bibfnamefont {S.}~\bibnamefont {Stepney}},\ and\ \bibinfo {author} {\bibfnamefont {M.~A.}\ \bibnamefont {Trefzer}},\ }\bibfield  {title} {\bibinfo {title} {Evolving carbon nanotube reservoir computers},\ }in\ \href@noop {} {\emph {\bibinfo {booktitle} {Unconventional Computation and Natural Computation}}}\ (\bibinfo  {publisher} {Springer International Publishing},\ \bibinfo {year} {2016})\ pp.\ \bibinfo {pages} {49--61}\BibitemShut {NoStop}%
\bibitem [{\citenamefont {Lee}\ \emph {et~al.}(2024)\citenamefont {Lee} \emph {et~al.}}]{lee2024task}%
  \BibitemOpen
  \bibfield  {author} {\bibinfo {author} {\bibfnamefont {O.}~\bibnamefont {Lee}} \emph {et~al.},\ }\bibfield  {title} {\bibinfo {title} {Task-adaptive physical reservoir computing},\ }\href@noop {} {\bibfield  {journal} {\bibinfo  {journal} {Nature Materials}\ }\textbf {\bibinfo {volume} {23}},\ \bibinfo {pages} {79} (\bibinfo {year} {2024})}\BibitemShut {NoStop}%
\bibitem [{\citenamefont {Kulkarni}\ and\ \citenamefont {Teuscher}(2012)}]{kulkarni2012memristor}%
  \BibitemOpen
  \bibfield  {author} {\bibinfo {author} {\bibfnamefont {M.~S.}\ \bibnamefont {Kulkarni}}\ and\ \bibinfo {author} {\bibfnamefont {C.}~\bibnamefont {Teuscher}},\ }\bibfield  {title} {\bibinfo {title} {Memristor-based reservoir computing},\ }in\ \href@noop {} {\emph {\bibinfo {booktitle} {Proceedings of the 2012 IEEE/ACM International Symposium on Nanoscale Architectures (NANOARCH)}}}\ (\bibinfo  {publisher} {Association for Computing Machinery},\ \bibinfo {address} {New York, NY, USA},\ \bibinfo {year} {2012})\ pp.\ \bibinfo {pages} {226--232}\BibitemShut {NoStop}%
\bibitem [{\citenamefont {Du}\ \emph {et~al.}(2017)\citenamefont {Du}, \citenamefont {Cai}, \citenamefont {Zidan}, \citenamefont {Ma}, \citenamefont {Lee},\ and\ \citenamefont {Lu}}]{du2017reservoir}%
  \BibitemOpen
  \bibfield  {author} {\bibinfo {author} {\bibfnamefont {C.}~\bibnamefont {Du}}, \bibinfo {author} {\bibfnamefont {F.}~\bibnamefont {Cai}}, \bibinfo {author} {\bibfnamefont {M.~A.}\ \bibnamefont {Zidan}}, \bibinfo {author} {\bibfnamefont {W.}~\bibnamefont {Ma}}, \bibinfo {author} {\bibfnamefont {S.~H.}\ \bibnamefont {Lee}},\ and\ \bibinfo {author} {\bibfnamefont {W.~D.}\ \bibnamefont {Lu}},\ }\bibfield  {title} {\bibinfo {title} {Reservoir computing using dynamic memristors for temporal information processing},\ }\href@noop {} {\bibfield  {journal} {\bibinfo  {journal} {Nature Communications}\ }\textbf {\bibinfo {volume} {8}},\ \bibinfo {pages} {2204} (\bibinfo {year} {2017})}\BibitemShut {NoStop}%
\bibitem [{\citenamefont {Appeltant}\ \emph {et~al.}(2011)\citenamefont {Appeltant} \emph {et~al.}}]{appeltant2011information}%
  \BibitemOpen
  \bibfield  {author} {\bibinfo {author} {\bibfnamefont {L.}~\bibnamefont {Appeltant}} \emph {et~al.},\ }\bibfield  {title} {\bibinfo {title} {Information processing using a single dynamical node as complex system},\ }\href@noop {} {\bibfield  {journal} {\bibinfo  {journal} {Nature Communications}\ }\textbf {\bibinfo {volume} {2}},\ \bibinfo {pages} {468} (\bibinfo {year} {2011})}\BibitemShut {NoStop}%
\bibitem [{\citenamefont {Balke}\ \emph {et~al.}(2008)\citenamefont {Balke}, \citenamefont {Granzow},\ and\ \citenamefont {R{\"o}del}}]{balke2008current}%
  \BibitemOpen
  \bibfield  {author} {\bibinfo {author} {\bibfnamefont {N.}~\bibnamefont {Balke}}, \bibinfo {author} {\bibfnamefont {T.}~\bibnamefont {Granzow}},\ and\ \bibinfo {author} {\bibfnamefont {J.}~\bibnamefont {R{\"o}del}},\ }\bibfield  {title} {\bibinfo {title} {Current-voltage characteristics for lead zirconate titanate bulk ceramics},\ }\href@noop {} {\bibfield  {journal} {\bibinfo  {journal} {Journal of Applied Physics}\ }\textbf {\bibinfo {volume} {104}},\ \bibinfo {pages} {054120} (\bibinfo {year} {2008})}\BibitemShut {NoStop}%
\bibitem [{\citenamefont {Rietman}\ \emph {et~al.}(2022)\citenamefont {Rietman}, \citenamefont {Schumm}, \citenamefont {Salik}, \citenamefont {Askenazi},\ and\ \citenamefont {Siegelmann}}]{rietman2022machine}%
  \BibitemOpen
  \bibfield  {author} {\bibinfo {author} {\bibfnamefont {E.}~\bibnamefont {Rietman}}, \bibinfo {author} {\bibfnamefont {L.}~\bibnamefont {Schumm}}, \bibinfo {author} {\bibfnamefont {A.}~\bibnamefont {Salik}}, \bibinfo {author} {\bibfnamefont {M.}~\bibnamefont {Askenazi}},\ and\ \bibinfo {author} {\bibfnamefont {H.}~\bibnamefont {Siegelmann}},\ }\bibfield  {title} {\bibinfo {title} {Machine learning with quantum matter: An example using lead zirconate titanate},\ }\href@noop {} {\bibfield  {journal} {\bibinfo  {journal} {Quantum Reports}\ }\textbf {\bibinfo {volume} {4}},\ \bibinfo {pages} {418} (\bibinfo {year} {2022})}\BibitemShut {NoStop}%
\bibitem [{\citenamefont {LeCun}(1998)}]{lecun1998mnist}%
  \BibitemOpen
  \bibfield  {author} {\bibinfo {author} {\bibfnamefont {Y.}~\bibnamefont {LeCun}},\ }\href {http://yann.lecun.com/exdb/mnist/} {\bibinfo {title} {The {MNIST} database of handwritten digits}} (\bibinfo {year} {1998})\BibitemShut {NoStop}%
\bibitem [{\citenamefont {Becker}\ \emph {et~al.}(2024)\citenamefont {Becker}, \citenamefont {Vielhaben}, \citenamefont {Ackermann}, \citenamefont {M{\"u}ller}, \citenamefont {Lapuschkin},\ and\ \citenamefont {Samek}}]{becker2024audiomnist}%
  \BibitemOpen
  \bibfield  {author} {\bibinfo {author} {\bibfnamefont {S.}~\bibnamefont {Becker}}, \bibinfo {author} {\bibfnamefont {J.}~\bibnamefont {Vielhaben}}, \bibinfo {author} {\bibfnamefont {M.}~\bibnamefont {Ackermann}}, \bibinfo {author} {\bibfnamefont {K.-R.}\ \bibnamefont {M{\"u}ller}}, \bibinfo {author} {\bibfnamefont {S.}~\bibnamefont {Lapuschkin}},\ and\ \bibinfo {author} {\bibfnamefont {W.}~\bibnamefont {Samek}},\ }\bibfield  {title} {\bibinfo {title} {{AudioMNIST}: Exploring explainable artificial intelligence for audio analysis on a simple benchmark},\ }\href@noop {} {\bibfield  {journal} {\bibinfo  {journal} {Journal of the Franklin Institute}\ }\textbf {\bibinfo {volume} {361}},\ \bibinfo {pages} {418} (\bibinfo {year} {2024})}\BibitemShut {NoStop}%
\bibitem [{\citenamefont {Lee}\ and\ \citenamefont {Mochizuki}(2023)}]{lee2023handwritten}%
  \BibitemOpen
  \bibfield  {author} {\bibinfo {author} {\bibfnamefont {M.-K.}\ \bibnamefont {Lee}}\ and\ \bibinfo {author} {\bibfnamefont {M.}~\bibnamefont {Mochizuki}},\ }\bibfield  {title} {\bibinfo {title} {Handwritten digit recognition by spin waves in a skyrmion reservoir},\ }\href@noop {} {\bibfield  {journal} {\bibinfo  {journal} {Scientific Reports}\ }\textbf {\bibinfo {volume} {13}},\ \bibinfo {pages} {19423} (\bibinfo {year} {2023})}\BibitemShut {NoStop}%
\bibitem [{\citenamefont {Schaetti}\ \emph {et~al.}(2016)\citenamefont {Schaetti}, \citenamefont {Salomon},\ and\ \citenamefont {Couturier}}]{schaetti2016echo}%
  \BibitemOpen
  \bibfield  {author} {\bibinfo {author} {\bibfnamefont {N.}~\bibnamefont {Schaetti}}, \bibinfo {author} {\bibfnamefont {M.}~\bibnamefont {Salomon}},\ and\ \bibinfo {author} {\bibfnamefont {R.}~\bibnamefont {Couturier}},\ }\bibfield  {title} {\bibinfo {title} {Echo state networks-based reservoir computing for {MNIST} handwritten digits recognition},\ }in\ \href@noop {} {\emph {\bibinfo {booktitle} {2016 IEEE International Conference on Computational Science and Engineering (CSE) and IEEE International Conference on Embedded and Ubiquitous Computing (EUC) and 15th International Symposium on Distributed Computing and Applications for Business Engineering (DCABES)}}}\ (\bibinfo  {publisher} {IEEE},\ \bibinfo {year} {2016})\ pp.\ \bibinfo {pages} {484--491}\BibitemShut {NoStop}%
\bibitem [{\citenamefont {Moon}\ \emph {et~al.}(2001)\citenamefont {Moon}, \citenamefont {Jagadish}, \citenamefont {Faloutsos},\ and\ \citenamefont {Saltz}}]{moon2001analysis}%
  \BibitemOpen
  \bibfield  {author} {\bibinfo {author} {\bibfnamefont {B.}~\bibnamefont {Moon}}, \bibinfo {author} {\bibfnamefont {H.~V.}\ \bibnamefont {Jagadish}}, \bibinfo {author} {\bibfnamefont {C.}~\bibnamefont {Faloutsos}},\ and\ \bibinfo {author} {\bibfnamefont {J.~H.}\ \bibnamefont {Saltz}},\ }\bibfield  {title} {\bibinfo {title} {Analysis of the clustering properties of the {Hilbert} space-filling curve},\ }\href@noop {} {\bibfield  {journal} {\bibinfo  {journal} {IEEE Transactions on Knowledge and Data Engineering}\ }\textbf {\bibinfo {volume} {13}},\ \bibinfo {pages} {124} (\bibinfo {year} {2001})}\BibitemShut {NoStop}%
\bibitem [{\citenamefont {Amaral}\ \emph {et~al.}(2022)\citenamefont {Amaral}, \citenamefont {Chester}, \citenamefont {Fang},\ and\ \citenamefont {Irwin}}]{amaral2022exploiting}%
  \BibitemOpen
  \bibfield  {author} {\bibinfo {author} {\bibfnamefont {M.}~\bibnamefont {Amaral}}, \bibinfo {author} {\bibfnamefont {D.}~\bibnamefont {Chester}}, \bibinfo {author} {\bibfnamefont {F.}~\bibnamefont {Fang}},\ and\ \bibinfo {author} {\bibfnamefont {K.}~\bibnamefont {Irwin}},\ }\bibfield  {title} {\bibinfo {title} {Exploiting anyonic behavior of quasicrystals for topological quantum computing},\ }\href@noop {} {\bibfield  {journal} {\bibinfo  {journal} {Symmetry}\ }\textbf {\bibinfo {volume} {14}},\ \bibinfo {pages} {1780} (\bibinfo {year} {2022})}\BibitemShut {NoStop}%
\bibitem [{\citenamefont {Dorini}\ \emph {et~al.}(2021)\citenamefont {Dorini}, \citenamefont {Brix}, \citenamefont {Chatelier}, \citenamefont {Kokalj},\ and\ \citenamefont {Gaudry}}]{dorini2021two}%
  \BibitemOpen
  \bibfield  {author} {\bibinfo {author} {\bibfnamefont {T.~T.}\ \bibnamefont {Dorini}}, \bibinfo {author} {\bibfnamefont {F.}~\bibnamefont {Brix}}, \bibinfo {author} {\bibfnamefont {C.}~\bibnamefont {Chatelier}}, \bibinfo {author} {\bibfnamefont {A.}~\bibnamefont {Kokalj}},\ and\ \bibinfo {author} {\bibfnamefont {E.}~\bibnamefont {Gaudry}},\ }\bibfield  {title} {\bibinfo {title} {Two-dimensional oxide quasicrystal approximants with tunable electronic and magnetic properties},\ }\href@noop {} {\bibfield  {journal} {\bibinfo  {journal} {Nanoscale}\ }\textbf {\bibinfo {volume} {13}},\ \bibinfo {pages} {10771} (\bibinfo {year} {2021})}\BibitemShut {NoStop}%
\bibitem [{\citenamefont {Forster}\ \emph {et~al.}(2013)\citenamefont {Forster}, \citenamefont {Meinel}, \citenamefont {Hammer}, \citenamefont {Troutmann},\ and\ \citenamefont {Widdra}}]{forster2013quasicrystalline}%
  \BibitemOpen
  \bibfield  {author} {\bibinfo {author} {\bibfnamefont {S.}~\bibnamefont {Forster}}, \bibinfo {author} {\bibfnamefont {K.}~\bibnamefont {Meinel}}, \bibinfo {author} {\bibfnamefont {R.}~\bibnamefont {Hammer}}, \bibinfo {author} {\bibfnamefont {M.}~\bibnamefont {Troutmann}},\ and\ \bibinfo {author} {\bibfnamefont {W.}~\bibnamefont {Widdra}},\ }\bibfield  {title} {\bibinfo {title} {Quasicrystalline structure formation in classical crystalline thin film system},\ }\href@noop {} {\bibfield  {journal} {\bibinfo  {journal} {Nature}\ }\textbf {\bibinfo {volume} {502}},\ \bibinfo {pages} {215} (\bibinfo {year} {2013})}\BibitemShut {NoStop}%
\bibitem [{\citenamefont {Huran}\ \emph {et~al.}(2021)\citenamefont {Huran}, \citenamefont {Wang},\ and\ \citenamefont {Marques}}]{huran2021two}%
  \BibitemOpen
  \bibfield  {author} {\bibinfo {author} {\bibfnamefont {A.~W.}\ \bibnamefont {Huran}}, \bibinfo {author} {\bibfnamefont {H.-C.}\ \bibnamefont {Wang}},\ and\ \bibinfo {author} {\bibfnamefont {M.~A.~L.}\ \bibnamefont {Marques}},\ }\bibfield  {title} {\bibinfo {title} {Two-dimensional binary metal-oxide quasicrystal approximants},\ }\href@noop {} {\bibfield  {journal} {\bibinfo  {journal} {2D Materials}\ }\textbf {\bibinfo {volume} {8}},\ \bibinfo {pages} {045002} (\bibinfo {year} {2021})}\BibitemShut {NoStop}%
\bibitem [{\citenamefont {Krupski}\ \emph {et~al.}(2015)\citenamefont {Krupski}, \citenamefont {Moors}, \citenamefont {Jozwia}, \citenamefont {Kobiela},\ and\ \citenamefont {Krupski}}]{krupski2015structure}%
  \BibitemOpen
  \bibfield  {author} {\bibinfo {author} {\bibfnamefont {K.}~\bibnamefont {Krupski}}, \bibinfo {author} {\bibfnamefont {M.}~\bibnamefont {Moors}}, \bibinfo {author} {\bibfnamefont {P.}~\bibnamefont {Jozwia}}, \bibinfo {author} {\bibfnamefont {T.}~\bibnamefont {Kobiela}},\ and\ \bibinfo {author} {\bibfnamefont {A.}~\bibnamefont {Krupski}},\ }\bibfield  {title} {\bibinfo {title} {Structure determination of {Au} on {Pt}(111) surface: {LEED}, {STM}, and {DFT} study},\ }\href@noop {} {\bibfield  {journal} {\bibinfo  {journal} {Materials}\ }\textbf {\bibinfo {volume} {8}},\ \bibinfo {pages} {2935} (\bibinfo {year} {2015})}\BibitemShut {NoStop}%
\bibitem [{\citenamefont {Yoneda}\ \emph {et~al.}(1998)\citenamefont {Yoneda}, \citenamefont {Okabe}, \citenamefont {Sakaue},\ and\ \citenamefont {Terauchi}}]{yoneda1998growth}%
  \BibitemOpen
  \bibfield  {author} {\bibinfo {author} {\bibfnamefont {Y.}~\bibnamefont {Yoneda}}, \bibinfo {author} {\bibfnamefont {T.}~\bibnamefont {Okabe}}, \bibinfo {author} {\bibfnamefont {T.}~\bibnamefont {Sakaue}},\ and\ \bibinfo {author} {\bibfnamefont {H.}~\bibnamefont {Terauchi}},\ }\bibfield  {title} {\bibinfo {title} {Growth of ultrathin {BaTiO}$_3$ films on {SrTiO}$_3$ and {MgO} substrates},\ }\href@noop {} {\bibfield  {journal} {\bibinfo  {journal} {Surface Science}\ }\textbf {\bibinfo {volume} {410}},\ \bibinfo {pages} {62} (\bibinfo {year} {1998})}\BibitemShut {NoStop}%
\bibitem [{\citenamefont {Yu}\ and\ \citenamefont {Liu}(2020)}]{yu2020piezoelectricity}%
  \BibitemOpen
  \bibfield  {author} {\bibinfo {author} {\bibfnamefont {J.}~\bibnamefont {Yu}}\ and\ \bibinfo {author} {\bibfnamefont {C.-X.}\ \bibnamefont {Liu}},\ }\bibfield  {title} {\bibinfo {title} {Piezoelectricity and topological quantum phase transitions in two-dimensional spin-orbit coupled crystals with time-reversal symmetry},\ }\href@noop {} {\bibfield  {journal} {\bibinfo  {journal} {Nature Communications}\ }\textbf {\bibinfo {volume} {11}},\ \bibinfo {pages} {2290} (\bibinfo {year} {2020})}\BibitemShut {NoStop}%
\bibitem [{\citenamefont {Zali}\ \emph {et~al.}(2014)\citenamefont {Zali}, \citenamefont {Mahmood}, \citenamefont {Mahamad}, \citenamefont {Foo},\ and\ \citenamefont {Murshidi}}]{zali2014xray}%
  \BibitemOpen
  \bibfield  {author} {\bibinfo {author} {\bibfnamefont {N.~M.}\ \bibnamefont {Zali}}, \bibinfo {author} {\bibfnamefont {C.~S.}\ \bibnamefont {Mahmood}}, \bibinfo {author} {\bibfnamefont {S.~M.}\ \bibnamefont {Mahamad}}, \bibinfo {author} {\bibfnamefont {C.~T.}\ \bibnamefont {Foo}},\ and\ \bibinfo {author} {\bibfnamefont {J.~A.}\ \bibnamefont {Murshidi}},\ }\bibfield  {title} {\bibinfo {title} {X-ray diffraction study of crystalline barium titanate ceramics},\ }in\ \href@noop {} {\emph {\bibinfo {booktitle} {Advances in Nuclear Research and Energy Development, AIP Conference Proceedings}}},\ Vol.\ \bibinfo {volume} {1584}\ (\bibinfo {year} {2014})\ pp.\ \bibinfo {pages} {160--163}\BibitemShut {NoStop}%
\end{thebibliography}%

\end{document}